\documentclass[a4paper, 10pt, conference]{ieeeconf}      

\IEEEoverridecommandlockouts                              

\usepackage[binary-units=true]{siunitx}
\usepackage{pgf}
\usepackage{tikz}
\usepackage{amssymb}  
\usepackage{graphics} 
\usepackage{epsfig} 
\usepackage{mathptmx} 
\usepackage{times} 
\usepackage{amsmath} 
\usepackage{mathrsfs}
\usepackage{booktabs}
\usepackage{bm}

\usepackage{hyperref}

\usepackage{pgfplots}
\pgfplotsset{compat=newest}
\usetikzlibrary{arrows,positioning,fit,shapes,calc, shapes.geometric}
\usetikzlibrary{plotmarks}
\usetikzlibrary{arrows.meta}
\usetikzlibrary{plotmarks}
\usepgfplotslibrary{patchplots}
\usepackage{grffile}
\usepackage{amsmath}

\newcommand{\etal}[1]{\mbox{{#1} \textit{et al.}}}

\usepackage[top=1.46in, bottom=0.77in,left=0.75in, right=0.52in ]{geometry}

\newcommand\copyrighttext{%
	\scriptsize \copyright~2019 IEEE. Personal use of this material is permitted. Permission from IEEE must be obtained for all other uses, in any current or future media, including reprinting/republishing this material for advertising or promotional purposes, creating new collective works, for resale or redistribution to servers or lists, or reuse of any copyrighted component of this work in other works.}%
\newcommand\copyrightnotice{%
\begin{tikzpicture}[remember picture,overlay]
\node[anchor=south,yshift=10pt,xshift=0.25cm] at (current page.south) {{\parbox{\dimexpr\textwidth-\fboxsep-\fboxrule\relax}{\copyrighttext}}};
\end{tikzpicture}%
}

\title{\LARGE \bf
DeepLocalization:\\ Landmark-based Self-Localization with Deep Neural Networks
}

\author{Nico Engel, Stefan Hoermann, Markus Horn, Vasileios Belagiannis and Klaus Dietmayer%
\thanks{The authors are with Institute of Measurement, Control and \mbox{Microtechnology}, Ulm University, 89081 Ulm, Germany.
       \mbox{E-Mail: {\tt\small \{firstname.lastname\}@uni-ulm.de}}. 
       Project page: \footnotesize \mbox{\url{https://www.uni-ulm.de/in/mrm/deeplocalization}}}%
}

\begin{document}

\maketitle
\copyrightnotice
\thispagestyle{empty}
\pagestyle{empty}

\begin{abstract}\label{sec:00_abstract}
	We address the problem of vehicle self-localization from multi-modal sensor information and a reference map. The map is generated off-line by extracting landmarks from the vehicle's field of view, while the measurements are collected similarly on the fly.
	Our goal is to determine the autonomous vehicle's pose from the landmark measurements and map landmarks.
	To learn this mapping, we propose  \mbox{DeepLocalization}, a deep neural network that regresses the vehicle's translation and rotation parameters from unordered and dynamic input landmarks. 	
	The proposed network architecture is robust to changes of the dynamic environment and can cope with a small number of extracted landmarks. During the training process we rely on synthetically generated ground-truth. 	
	In our experiments, we evaluate two inference approaches in real-world scenarios. We show that DeepLocalization can be combined with regular GPS signals and filtering algorithms such as the extended Kalman filter. Our approach achieves state-of-the-art accuracy and is about ten times faster than the related work.	
\end{abstract}

\section{Introduction}\label{sec:01_introduction}

Self-localization is the task of determining the vehicle's position and orientation (pose) within a coordinate system, based on sensor information \cite{thrun2005probabilistic}. In autonomous driving, the precise vehicle position is essential for path planning~\cite{kunz2015autonomous}. Furthermore, it increases the vehicle's safety and reliability  by incorporating prior information from a digital map for the environment perception~\cite{gies2018environment}.

The required localization accuracy is, of course, subject to the driving environment. In structured and organized environments, e.g. freeways, the localization accuracy can be as low as a couple of meters under certain conditions. Regular global navigation satellite systems (GNSS), such as GPS, usually provide sufficient results for these tasks~\cite{mohamed1999adaptive}. 
However, more complex and unstructured scenarios in urban and rural areas demand an extensive environment perception with a precise localization, requiring accuracies in the range of $\SI[mode=text]{0.2}{\meter} - \SI[mode=text]{0.5}{\meter}$. Achieving this precision is possible with a dGPS-System that relies on correction data from a network of ground-based reference stations. However, this is not a sustainable solution for every autonomous driving vehicle due to the high cost of the system. In addition, all GNSS systems heavily rely on satellite coverage, which can be poor in cities where the clear sight to the satellites is concealed by tall buildings~\cite{kos2010effects}. The most adopted solution in localization is to combine multiple sensors, e.g.~radar, camera, lidar and GPS, with a high-precision map to estimate the vehicle's pose~\cite{cummins2008fab, ziegler2014video, brubaker2016map}.

Classical probabilistic localization approaches have shown great results in unstructured environments enabling the successful deployment of autonomous vehicles in many cities around the world using either a grid-based representation of the environment~\cite{wiest2014localization} or a landmark-based map. In the latter case, the map contains static and easily recognizable objects~\cite{deusch2015labeled, stubler2017consistency}. Storing objects in a landmarks-based map requires small amount of memory compared to the grid-based maps, where the area is discretized into small cells. Moreover, landmark-based maps are easier to maintain and update, because landmarks can be added or removed without much effort.

\begin{figure}[t]
    \centering
    \input{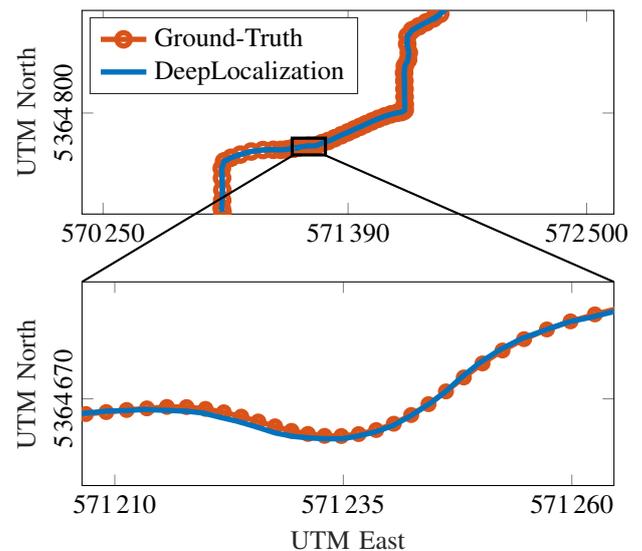}
    \caption{Results of DeepLocalization. The plot on the top shows an extract from our evaluation sequence, where the red line is the ground-truth trajectory and the blue line the output from our network in combination with an EKF. The bottom plot shows an enlarged section of the top trajectory.}
    \label{fig:overview_plot}
\end{figure}

In this work, we present an approach for self-localization based on landmarks. By relying on multi-modal sensor information, we extract a set of landmarks from the vehicle's field of view and build a map off-line. During localization, the autonomous vehicle extracts landmarks on the fly that are matched with the map landmarks and, eventually, estimates the vehicle's pose. Note that there are map landmarks which will not be visible after building the map due to the dynamic environment (e.g different lighting and weather conditions, dynamic objects and infrastructure). Furthermore, estimating the vehicle's absolute pose in a global coordinate system is highly ambiguous and thus we rely on the vehicle's pose from the previous time step to estimate the current pose. On top of the environment challenges, we also have to deal with input of dynamic size since the number of landmarks in the vehicle's field of view varies over time. As a result, matching the measured landmarks that are extracted in real-time with the map landmarks is not straightforward. To address these issues, we introduce a deep neural network that learns to match the measured landmarks with the map landmarks based on synthetically generated ground-truth.

Traditional convolutional neural networks are not well-suited for our problem, because they require the input data to be structured. Moreover, transforming the dynamic input into an ordered representation is not trivial. We derive our motivation from the latest network architectures for unstructured 3D data processing. The recently introduced \mbox{PointNet} architecture~\cite{qi2017pointnet} copes with unordered point lists. The main idea of PointNet is to create a signature of the dynamic input list, i.e. the 3D point cloud, by combining a multi-layer perceptron (MLP) with the max-pooling operation.

We propose a neural network architecture that receives input from two dynamic lists to regress the vehicle's pose. The measured landmarks compose one dynamic list and the map landmarks the other one. Our architecture does not explicitly model the matching process. Instead, it is a vehicle's pose regressor with measurements and map landmarks as input. Moreover, the pose is not expressed in the global coordinate system as discussed above. Our networks regress the vehicle's pose w.r.t previous pose, i.e.  a pose offset. During inference, our only requirement is the availability of at least a single GPS measurement to initiate our pipeline.  We present two inference algorithms to show that our approach improves the GPS measurements and is also compatible with filtering algorithms, such as the extended Kalman filter.

In our experiments, we study the performance of our approach on our own dataset which was recorded in Ulm-Lehr, Germany. First, we perform a number of synthetic experiments to show the generalization capabilities of our network. Then, we compare our results with prior work to show better localization accuracy and faster inference times in the range of $2\,\si{\milli\second}$. We call our approach \textit{\mbox{DeepLocalization}}. Some of our results are shown in \mbox{Fig. \ref{fig:overview_plot}}.

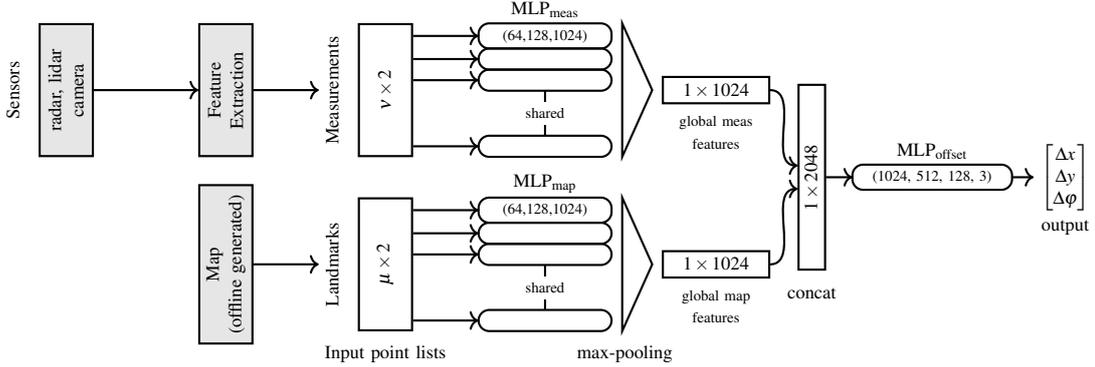
\begin{figure*}[t]
    \centering
    \begin{tikzpicture}[thick, scale=0.7, every node/.style={scale=0.7}]

    \node[ rectangle, draw, fill=black!10, rotate=90, align=center, inner sep=0.35cm, minimum width=2.5cm, minimum height=1cm, text width=2.5cm, inner sep=0pt] at (-6cm, 0cm) (sensors) { radar, lidar \\  camera};
    \node[ rectangle, draw, fill=black!10,rotate=90, align=center, inner sep=0.35cm, minimum width=2.5cm, minimum height=1cm, text width=2.5cm, inner sep=0pt] at (-3cm, 0cm) (feature_extraction) { Feature Extraction };
    \node[ rectangle, draw, fill=black!10, rotate=90, align=center, inner sep=0.35cm, minimum width=3cm, minimum height=1cm, text width=3cm, inner sep=0pt] at (-3cm, -3.3cm) (hdmap) { Map \\ (offline generated)};
    \node[rectangle, draw, rotate=90, inner sep=0.35cm, minimum width=2.5cm, minimum height=1cm, inner sep=0pt] (meas) {$\nu \times 2$};
    \node[rectangle, draw, rotate=90, inner sep=0.35cm, minimum width=2.5cm, minimum height=1cm, xshift=-3.3cm, inner sep= 0pt] (map) {$\mu \times 2$};

    \node [left of= meas, rotate=90] (text_feat) {Measurements};
    \node [left of= map, rotate=90] (text_landmarks) {Landmarks};
    \node [left of= sensors, rotate=90] (text_sensors) {Sensors};

    \node[rectangle, draw, rounded corners=4pt, minimum width = 2.5cm, minimum height=0.4cm, anchor=north, at=(meas.east), xshift=3cm, outer sep=0pt] (mlp_meas1) {\footnotesize (64,128,1024)};
    \node[rectangle, draw, rounded corners=4pt, minimum width = 2.5cm, minimum height=0.4cm, anchor=north, at=(mlp_meas1.south), outer sep=0pt] (mlp_meas2) {};  
    \node[rectangle, draw, rounded corners=4pt, minimum width = 2.5cm, minimum height=0.4cm, anchor=north, at=(mlp_meas2.south), outer sep=0pt] (mlp_meas3) {}; 

    \node[rectangle, draw, rounded corners=4pt, minimum width = 2.5cm, minimum height=0.4cm, anchor=south, at=(meas.west), xshift=3cm, outer sep=0pt] (mlp_measn) {};

    \draw (mlp_meas3.south) --(mlp_measn.north)  node[midway, fill=white] {\footnotesize shared} ;
    \node [rectangle, above of=mlp_meas1, yshift=-0.5cm, outer sep=0pt] {$\text{MLP}_{\text{meas}}$};

    \node[rectangle, draw, rounded corners=4pt, minimum width = 2.5cm, minimum height=0.4cm, anchor=north, at=(map.east), xshift=3cm, outer sep=0pt] (mlp_map1) {\footnotesize (64,128,1024)};
    \node[rectangle, draw, rounded corners=4pt, minimum width = 2.5cm, minimum height=0.4cm, anchor=north, at=(mlp_map1.south), outer sep=0pt] (mlp_map2) {};  
    \node[rectangle, draw, rounded corners=4pt, minimum width = 2.5cm, minimum height=0.4cm, anchor=north, at=(mlp_map2.south), outer sep=0pt] (mlp_map3) {}; 

    \node [rectangle, above of=mlp_map1, yshift=-0.5cm, outer sep=0pt] {$\text{MLP}_{\text{map}}$};

    \node[rectangle, draw, rounded corners=4pt, minimum width = 2.5cm, minimum height=0.4cm, anchor=south, at=(map.west), xshift=3cm, outer sep=0pt] (mlp_mapn) {};

    \draw (mlp_map3.south) --(mlp_mapn.north)  node[midway, fill=white] {\footnotesize shared} ;

    \draw [->] (sensors.south) -- (feature_extraction.north);
    \draw [->] (feature_extraction.south) -- (text_feat.north);
    \draw [->] (hdmap.south) -- (text_landmarks.north);

    \draw [->] ($(mlp_meas1) - (2.48cm, 0)$) -- (mlp_meas1); 
    \draw [->] ($(mlp_meas2) - (2.48cm, 0)$) -- (mlp_meas2); 
    \draw [->] ($(mlp_meas3) - (2.48cm, 0)$) -- (mlp_meas3); 
    \draw [->] ($(mlp_measn) - (2.48cm, 0)$) -- (mlp_measn); 

    \draw [->] ($(mlp_map1) - (2.48cm, 0)$) -- (mlp_map1); 
    \draw [->] ($(mlp_map2) - (2.48cm, 0)$) -- (mlp_map2); 
    \draw [->] ($(mlp_map3) - (2.48cm, 0)$) -- (mlp_map3); 
    \draw [->] ($(mlp_mapn) - (2.48cm, 0)$) -- (mlp_mapn); 

    \draw ($(mlp_meas1.north east) + (0.2cm,0)$) -- ($(mlp_measn.south east) + (0.2cm,0)$) -- ($(meas.south) + (4.5cm,0)$) coordinate(max_meas) -- cycle;

    \draw ($(mlp_map1.north east) + (0.2cm,0)$) -- ($(mlp_mapn.south east) + (0.2cm,0)$) -- ($(map.south) + (4.5cm,0)$) coordinate(max_map) -- cycle;

    \node [rectangle,xshift=0.2cm, minimum width=2cm, draw, right of=max_meas] (feature_meas) {$1\times1024$};
    \node [rectangle,xshift=0.2cm, minimum width=2cm, draw, right of=max_map] (feature_map) {$1\times1024$};

    \node[below of=feature_meas,align=center, yshift=0.2cm] {\footnotesize global meas \\ \footnotesize features};
    \node[below of=feature_map,align=center, yshift=0.2cm] {\footnotesize global map \\ \footnotesize features};

     \node [xshift=3cm, draw, minimum width=3.5cm, rotate=90] (concat) at ($(max_meas)!0.5!(max_map)$) {$1\times2048$};
     \node [below of=concat, yshift=-1.2cm] {concat};

     \draw [->] (feature_meas.east) to [out=0, in=180] (concat.50);
     \draw [->] (feature_map.east) to [out=0, in=180] (concat.130);

    \node[rectangle, draw, rounded corners=4pt, minimum width = 3cm, minimum height=0.4cm, at=(concat.south), xshift=2cm, outer sep=0pt] (mlp_concat) {\footnotesize (1024, 512, 128, 3)};
    \node [rectangle, above of=mlp_concat, yshift=-0.5cm, outer sep=0pt] {$\text{MLP}_{\text{offset}}$};
    \draw[->] (concat) to (mlp_concat);



    \node[right of=mlp_concat, xshift=1.5cm] (output_y) {$\begin{bmatrix}
        \Delta x \\
        \Delta y \\
        \Delta \varphi    
    \end{bmatrix}$};

    \draw [->] (mlp_concat.east) to (output_y.west);

    \node[yshift=-5cm] {Input point lists};
    \node[yshift=-5cm, xshift=4.5cm] {max-pooling};
    \node[yshift=-2.6cm, xshift=12.78cm] {output};

\end{tikzpicture}
    \caption{Network architecture. The raw measurements from the sensors are pre-processed using feature extraction algorithms. The input points are then transformed into higher dimensional space using independent multi-layer perceptron networks for the measurements and map landmarks. The parameters are shared for each input type. A max-pooling operation is performed on the transformed input points resulting in two feature vectors describing the global structure of each point list. The two feature vectors are concatenated and used as input for another multi-layer perceptron that estimates a transformation to correct an initial or previous localization result.}
    \label{fig:network_architecture}
\end{figure*}

\section{Related Work}\label{sec:02_related_work}

Self-Localization is a fundamental problem in robotics. Usually, the standard scenario is to estimate the robot's pose by accumulating noisy sensor measurements using filtering algorithms such as the extended Kalman filter (EKF) or  particle filter, i.e.~Monte-Carlo localization (MCL)~\mbox{\cite{thrun2005probabilistic, dellaert1999monte, thrun2001robust}.}
Later, self-localization has been raised to a vital task for autonomous driving vehicles too. Several approaches have been proposed in the past few years. Below, we discuss the methods that are related to ours as well as deep learning approaches.

\noindent \textbf{Classic Self-Localization.} A common way of categorizing the localization methods is based on the type of environment representation. The majority of the approaches rely on grid-based maps, i.e. a discretization of the environment into 2D or 3D cells. In that case, features of the environment are generated for every cell and saved in an off-line map. In the on-line phase, the same feature extraction algorithms are used for producing measurements. By aligning the off-line map with the measurements the vehicle's pose can be estimated. \etal{Levinson}~\cite{levinson2007map, levinson2010robust} proposed a probabilistic grid, where the remittance value of lidar measurements is modeled for each cell as its own Gaussian distribution. \etal{Wolcott}~\cite{wolcott2015fast}  proposed the Gaussian mixture maps in which each cell of the 2D grid contains Gaussian mixture models over the height (vertical structure), measured by multiple lidar scanners. The advantage over approaches which use the reflectivity of lidar scans, e.g.~\cite{levinson2010robust}, lies in the robust localization accuracy even in adverse weather conditions. Besides the grid-based environment representation, landmark-based maps are a popular choice in autonomous driving applications too. The maps usually contain sets of independent 2D or 3D points to represent distinct and recognizable objects. Traditionally, observed point clouds from different sensors are registered to the pre-built map using the iterative closest point algorithm~\cite{besl1992method} or similar variations~\cite{segal2009generalized}. Instead of searching for a single best solution for the registration problem, several approaches use Monte Carlo localization methods in combination with a digital map and sensor measurements to determine the vehicle's pose~\cite{stubler2017consistency, kummerle2008monte}.

\noindent \textbf{Deep Localization.} Deep learning approaches have shown promising results in regression task~\cite{belagiannis2015robust}, including localization. \etal{Kendall}~proposed PoseNet~\cite{kendall2015posenet}, a convolutional neural network that regresses the camera's $6$ degrees of freedom pose relative to the scene from a single RGB image. A similar approach has been introduced by \etal{Valada}~with VLocNet~\cite{valada2018deep}. The goal is to regress the global pose and simultaneously estimate the odometry between two frames. The network is based on residual neural networks that take two consecutive monocular images as input. DeepMapping~\cite{ding2018deepmapping} aligns point clouds to a global coordinate system, which is closer to our approach. The method regresses the sensor's pose using a deep neural network followed by a mapping network that models the structure of a scene using grid maps. Contrary to this method, our model directly learns the alignment transformation from the two dynamic inputs without requiring an intermediate step. 

Finally, it is important to note that we derive inspiration from PointNet~\cite{qi2017pointnet} to process unstructured data such as point clouds. PointNet uses sets of 3D points to learn  a global feature representation of the input which can be used for classification and segmentation. Although PointNet is not suited for processing input from multiple sources, it is related to our model. We propose an architecture within the context of localization that regresses the vehicle's pose from two dynamic input lists, the measurements and map landmarks.

\section{Method}\label{sec:03_method}

We consider multi-modal measurements from lidar, radar and camera sensors. For each time step, a set of landmarks is extracted from the raw data using pre-processing algorithms~\cite{stubler2017consistency}. All extracted landmarks from the vehicle's field of view  are registered to the same coordinate system. This is performed off-line and constitutes our map building process. In general, our map is comprised of 2D points, each resembling a static and recognizable object \cite{thrun2005probabilistic}. In the on-line phase, landmarks, which we refer to as \textit{\mbox{measurements}}, are extracted the same way. The goal is to localize the vehicle w.r.t to the \mbox{\textit{map landmarks}} by utilizing the measurements and the vehicle's initial position. 

It is important to mention that there exists no specific order in which the measurements are obtained. Furthermore, the number of landmarks is not known in advance and it varies over time. We present an approach that copes with dynamic and unordered input.

\subsection{Problem Definition}
We seek for mapping the measurements $\{\mathbf{z}_{1}, \dots ,\mathbf{z}_{\nu}\}$,~$\mathbf{z}_{\{\cdot\}} \in \mathbb{R}^2$ to the map landmarks $\{\mathbf{m}_{1}, \dots ,\mathbf{m}_{\mu}\}$,~$\mathbf{m}_{\{\cdot\}} \in \mathbb{R}^2$ to obtain the vehicle's pose \mbox{$[x,y,\varphi]$}. In our representation, the measurements and map landmarks are comprised of 2D points. Regressing directly the absolute position in a global coordinate system, such as the Universal Transverse Mercator (UTM) coordinate system, is highly ambiguous. Instead, we propose to infer the pose offset between the previous and current time step. We make use of the vehicle's pose from the previous (or initial) time step $t-1$ and the current measurements to predict the pose offset \mbox{$[\Delta x, \Delta y, \Delta\varphi]$} to the current pose. The position offset \mbox{$\Delta x$} and \mbox{$\Delta y$} can be interpreted as translation and the orientation offset \mbox{$\Delta\varphi$} as rotation.

We present a deep neural network architecture that receives the measurements of the current time step $t$ and the map landmarks from $t-1$ as input to predict the pose offset between the previous and current time step. The idea is to learn the relation between measurements and map landmarks. The proposed mapping is parameterized by $\theta$ and is given by:
\begin{equation}\label{eq:mapping_f}
f_{\theta}\left ( \left \{ \mathbf{z}_1, \dots, \mathbf{z}_\nu \right \}_{t}, \left \{ \mathbf{m}_1 , \dots , \mathbf{m}_\mu  \right \}_{t-1} \right)\rightarrow \left[ \Delta \hat{x}, \Delta \hat{y}, \Delta \hat{\varphi} \right].
\end{equation}
Note that we have 2-axis of translation and 1-axis of rotation in our problem, which is a common way of describing an agent's pose~\cite{thrun2005probabilistic}. Based on the output of $f_{\theta}$ and the previous vehicle's pose $\mathbf{p}_{t-1}$, we can obtain the current pose $\mathbf{p}_{t}$. This is described by:
\begin{equation}\label{eq:PoseVeh}
\mathbf{p}_{t} = \mathbf{p}_{t-1} + [\Delta \hat{x}, \Delta \hat{y}, \Delta \hat{\varphi}]^{T},
\end{equation}
where $\mathbf{p}_{t}, \mathbf{p}_{t-1} \in \mathbb{R}^{3}$.

\subsection{Objective}
The learning objective is to minimize the difference between the predicted $[\Delta \hat{x}, \Delta \hat{y}, \Delta \hat{\varphi}]$ and ground-truth pose offsets $[\Delta x, \Delta y, \Delta\varphi]$. We define the translation and rotation loss terms that are treated independently. The reason is the different magnitude of the  units, i.e. meters and degree, that require scaling the two terms for equal gradient contribution. At first, the translation loss is given by:
\begin{equation}\label{eq:objective_translation}
\mathrm{L}_{\mathrm{tran}} = \mathbb{E}\left[  (\Delta \hat{x}  - \Delta x)^{2}  \right] + \mathbb{E}\left[  (\Delta \hat{y}  - \Delta y)^{2}  \right].
\end{equation}
Secondly, the rotation loss is defined as: 
\begin{equation}\label{eq:objective_translation}
\mathrm{L}_{\mathrm{rot}} = \mathbb{E}\left[  (\Delta \hat{\varphi}  - \Delta \varphi)^{2}  \right].
\end{equation}
We found that manually tuning the weighting of the two terms is a tedious task. Furthermore, the weights can be sensitive to small modifications in the training data. Rather than choosing a data specific weighting strategy, we learn the weighting factors~\cite{kendall2018multi}. The complete loss function is defined as:
\begin{equation}\label{eq:geometric_loss}
\mathrm L= \mathrm L_{\mathrm{tran}} e^{-s_{\mathrm{tran}}} + s_{\mathrm{tran}} + \mathrm L_{\mathrm{rot}} e^{-s_{\mathrm{rot}}}+s_{\mathrm{rot}},
\end{equation}
where \mbox{$s_{\mathrm{tran}} = \log\sigma_{\mathrm{tran}}^{2}$} and \mbox{$s_{\mathrm{rot}} = \log\sigma_{\mathrm{rot}}^{2}$}. The terms $\sigma_{\mathrm{tran}}$ and  $\sigma_{\mathrm{rot}}$ denote the homoscedastic uncertainty, derived from Bayesian modeling~\cite{kendall2017uncertainties}. Homoscedastic uncertainty is a sub type of aleatoric uncertainty that is not dependent on the input data. It is therefore constant but varies between different tasks. Following~\cite{kendall2018multi}, homoscedastic uncertainty can be interpreted as task-dependent weighting, which can be learned by the network. In our formulation, the tasks correspond to the translation and rotation. Our loss function is similar to the geometric loss function \mbox{in \cite{kendall2017geometric}}, where the only differences are the degrees of freedom.

\subsection{Network Architecture}
DeepLocalization, our network architecture, is presented in Fig. \ref{fig:network_architecture}. The inputs of DeepLocalization are two independent point lists, namely the measurements and map landmarks.  For each of the two input lists we employ a separate MLP with three layers, denoted by $\text{MLP}_{\text{meas}}$ and $\text{MLP}_{\text{map}}$ in Fig. \ref{fig:network_architecture}, which transform the input to a higher dimensional representation  $\mathbb{R}^D$, where $D = 1024$. Note that the weights of $\text{MLP}_{\text{meas}}$ are shared for every measurement, similar for the map landmarks and $\text{MLP}_{\text{map}}$. The output of each MLP is $\nu \times 1024$ for the measurements and $\mu \times 1024$ for the map landmarks. At the end of $\text{MLP}_{\text{meas}}$ and $\text{MLP}_{\text{map}}$, a max-pooling operation follows to produce the global feature vector of $D$ elements. One can observe that the global feature vector does not depend on the number of input landmarks which makes the network invariant to the input size. In addition, the global feature vector is invariant to input permutations~\cite{qi2017pointnet}. Finally, the two global feature vectors are concatenated and fed to a third $\text{MLP}_{\text{offset}}$. The role of $\text{MLP}_{\text{offset}}$ is to find the correlation between the measurements and map landmarks, which now have been projected to the same dimensions. The output of $\text{MLP}_{\text{offset}}$ is the vehicle's pose offset.

It's important to note that the map landmarks are loaded only for a radius of $\SI[mode=text]{100}{\meter}$ around the vehicle's noisy position. During training, we do have the ground-truth information for the vehicle's location and, consequently we can load the corresponding map landmarks. During testing, we make use of the previous or initial vehicle's location to load the map landmarks. Although, the radius around the vehicle from which the landmarks are loaded is slightly delayed in time, we have found that it does not have any negative impact on the pose offset prediction.

\subsection{Training Process}\label{seq:train_pro}

The training of our model is performed with synthetic pose offsets. We randomly fetch from the training data the following: the vehicle's ground-truth pose in the UTM coordinate system $\mathbf{p}_{\mathrm{UTM}} = [x_{\mathrm{UTM}}, y_{\mathrm{UTM}}, \varphi_{\mathrm{UTM}}]$, the map landmarks \mbox{$\{\mathbf{m}_{1}, \dots, \mathbf{m}_{\mu} \}$} in the UTM coordinate system and the measurements \mbox{$\{\mathbf{z}_{1}, \dots, \mathbf{z}_{\nu} \}$} in the vehicle coordinate system. In addition, we randomly sample a pose offset \mbox{$[\Delta x, \Delta y, \Delta\varphi]_{\sigma_{x}, \sigma_{y}, \sigma_{\varphi}}$} from a uniform distribution on the \mbox{interval} $\Delta x \in [-\sigma_{x},\sigma_{x}]\,\si{\meter}$, $\Delta y \in [-\sigma_{y},\sigma_{y}]\,\si{\meter}$ and \mbox{$\Delta \varphi \in [-\sigma_{\varphi},\sigma_{\varphi}]^{\circ}$}. The pose offset composes the desired network output, i.e.~ground-truth, and is applied to the vehicle's pose at the UTM coordinate system such that it causes a shift. This is written as:
\begin{equation}\label{eq:pinit}
\hat{\mathbf{p}}_{\mathrm{UTM}} = \mathbf{p}_{\mathrm{UTM}} + [\Delta x, \Delta y, \Delta\varphi]^{T}_{\sigma_{x}, \sigma_{y}, \sigma_{\varphi}},
\end{equation}
where $\hat{\mathbf{p}}_{\mathrm{UTM}} =  [\hat{x}_{\mathrm{UTM}}, \hat{y}_{\mathrm{UTM}}, \hat{\varphi}_{\mathrm{UTM}}]$. Based on the shifted vehicle's pose in the UTM coordinate system, we define the homogeneous transformation matrix $H$ that transforms a measurement from the vehicle to the UTM coordinate system. This is given by:
\begin{equation}\label{eq:homogenous_transform}
H = \begin{bmatrix}
\cos (\hat{\varphi}_{\mathrm{UTM}}) & - \sin (\hat{\varphi}_{\mathrm{UTM}}) & \hat{x}_{\mathrm{UTM}} \\
\sin (\hat{\varphi}_{\mathrm{UTM}}) & \cos (\hat{\varphi}_{\mathrm{UTM}}) & \hat{y}_{\mathrm{UTM}} \\
0 & 0 & 1
\end{bmatrix}.
\end{equation}
We can use $H$ to transform all measurements to the UTM coordinate system. However, we perform the opposite by transforming the map landmarks to the vehicle coordinate system with the inverse transformation matrix $H^{-1}$ for numerical stability reasons. The range of values in the vehicle coordinate system is better suited for training a neural network. Moreover, we predict offsets, i.e.~translation and rotation, which are independent of the coordinate system. At this point, the map landmarks and the  measurements are in the vehicle coordinate system.

This procedure allows us, theoretically, to generate an infinite amount of training data. Moreover, the sampling order over time does not play any role in our optimization. Thus, we can sample training data at random time steps. 

\begin{figure}
    \centering
    \begin{tikzpicture}[thick, scale=0.77, every node/.style={scale=0.77}]

\node[rectangle,rotate=90,draw, minimum width=3cm, minimum height=1cm] (deeplocalization) {DeepLocalization};

\node[xshift = 1.5cm, inner sep=0pt] (corr_vec) {$\begin{bmatrix}
    \Delta x\\
    \Delta y\\
    \Delta \varphi
\end{bmatrix}$};

\node[xshift = 3cm, circle, draw] (transform) {+};

\node[xshift = 4cm] (pt) {$\mathbf{p}_{t}$};
\node[xshift = 5.3cm, rectangle, draw, minimum height=3cm] (ekf) {EKF};

\draw[->] ($(deeplocalization.north) + (-1cm,+0.7cm) $) node [xshift=-0.3cm, yshift=0.3cm, ] {Measurements} node [yshift=-0.3cm] {$\left \{ \mathbf{z}_1, \dots, \mathbf{z}_\nu \right \}_{t}$}-- ($(deeplocalization.north) + (0cm,+0.7cm)$);
\draw[->] ($(deeplocalization.north) + (-1cm,-0.7cm) $) node [xshift=-0.1cm, yshift=0.3cm] {Landmarks} node [xshift=-0.3cm, yshift=-0.3cm] {$\left \{ \mathbf{m}_1 , \dots , \mathbf{m}_\mu  \right \}_{t-1} $} -- ($(deeplocalization.north) + (0cm,-0.7cm)$);

\draw[->] (deeplocalization.south) -- (corr_vec.west);
\draw[->] (corr_vec.east) -- (transform.west);
\draw[->] (transform.east) -- (pt.west);
\draw[->] (pt.east) -- (ekf.west);

\node [xshift=7cm] (ekf_out) {$\mathbf{p}_{t}^{\mathrm{EKF}}$};

\draw[->] ($(transform.south) + (0, -1.5cm) $)  node [draw, rectangle, yshift=-0.45cm, text width=3cm, align=center] (init) {init/previous pose \\ $\mathbf{p}_{t-1}$} -- (transform.south);
\draw[->] (ekf.east) -- (ekf_out.west);
\draw[->, dotted] (ekf_out.south)  |- (init.east) node [xshift=1.2cm, yshift=0.25cm] {update};
\draw[->, dotted] (init.west)  -| ($(deeplocalization.west) - (2,0)$) node [xshift=1.8cm, yshift=-0.5cm] {load landmarks};

\end{tikzpicture}
    \caption{Overview of filter-based system architecture. The input to DeepLocalization are the measurements and the map landmarks that are transformed to the vehicle coordinate system using $H^{-1}$ with $\mathbf{p}_{t-1}$. The output of the network is applied to an initial or previous localization result. The corrected pose is the input measurement for an EKF with constant turn rate and velocity motion model. Therefore, the EKF predicts a smoothed pose based on the previous pose and the network output, which is then used for the next time step. }
    \label{fig:ekf}
\end{figure}
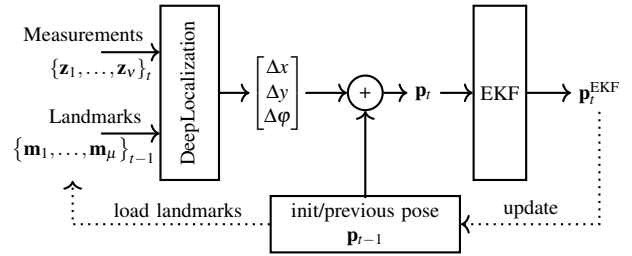

\section{Inference}
We present two inference approaches. First, the GPS-based inference is based on improving the GPS sensor measurement using our approach at every time step. Second, the filter-based inference combines our approach with an extended Kalman filter (EKF). The idea is to make use of the temporal information.

\subsection{GPS-based Inference}
The first inference algorithm relies on the GPS measurement for obtaining an initial but noisy vehicle pose. Here, we do not make use of the pose from the previous time step. Thus, we rewrite Eq.~\eqref{eq:PoseVeh} as:
\begin{equation}\label{eq:PoseVehGPS}
\mathbf{p}_{t} = \mathbf{p}_{\mathrm{GPS}} + [\Delta \hat{x}, \Delta \hat{y}, \Delta \hat{\varphi}]^{T},
\end{equation}
where $\mathbf{p}_{\mathrm{GPS}}$ is the noisy GPS vehicle pose in the UTM coordinate system. To obtain the pose offsets $\Delta \hat{x}$, $\Delta \hat{y}$ and $\Delta \hat{\varphi}$ from DeepLocalization, we follow the same pipeline as in the training process from Sec.~\ref{seq:train_pro}. The loaded map landmarks are transformed to the vehicle coordinate system based on $H^{-1}$, where the position and orientation from $\mathbf{p}_{\mathrm{GPS}}$ are used in $H$. The same process is followed at every time step to estimate $\mathbf{p}_{t}$.

\subsection{Filter-based Inference}
The filter-based algorithm requires an initial position, which can be retrieved from the GPS sensor. We rely on Eq.~\eqref{eq:PoseVeh} to predict the vehicle's position in the UTM coordinate system. At time step $t$, the measurements become available in the vehicle coordinate system. Moreover, the map landmarks are loaded based on the previous pose $\mathbf{p}_{t-1}$. The map landmarks are expressed in the vehicle coordinate system by applying the transformation $H^{-1}$, which is computed from the vehicle's previous pose $\mathbf{p}_{t-1}$, since $\mathbf{p}_{t}$ is not available. The measurements and map landmarks in the vehicle coordinate system are provided to DeepLocalization to regress the pose offset $[\Delta \hat{x}, \Delta \hat{y}, \Delta \hat{\varphi}]$. Finally, the new  pose $\mathbf{p}_{t}$ is calculated by applying the estimated offset to $\mathbf{p}_{t-1}$.

Nevertheless, our network does not take into consideration the time domain, which could help to avoid pose drifts. To address this issue, we additionally propose to combine our network with an EKF that has a constant turn rate and velocity (CTRV) motion model. The EKF takes the new vehicle pose $\mathbf{p}_{t}$, i.e.~Eq.~\eqref{eq:PoseVeh}, and refines it such that 
\begin{equation}
    \mathbf{p}_{t}^{\mathrm{EKF}}  = \mathrm{EKF}(\mathbf{p}_{t}),
\end{equation}
where $\mathrm{EKF}$ represents the filter. 
At the next time step, the pose $ \mathbf{p}_{t}^{\mathrm{EKF}}$ is used as $\mathbf{p}_{t-1}$ and the process is repeated with new measurements. An overview of the filter-based inference method is shown in Fig.~\ref{fig:ekf}.

\definecolor{landmark_rgb}{RGB}{235,72,104} 
\begin{figure}[t]
    \centering
    \includegraphics[width=0.45\textwidth]{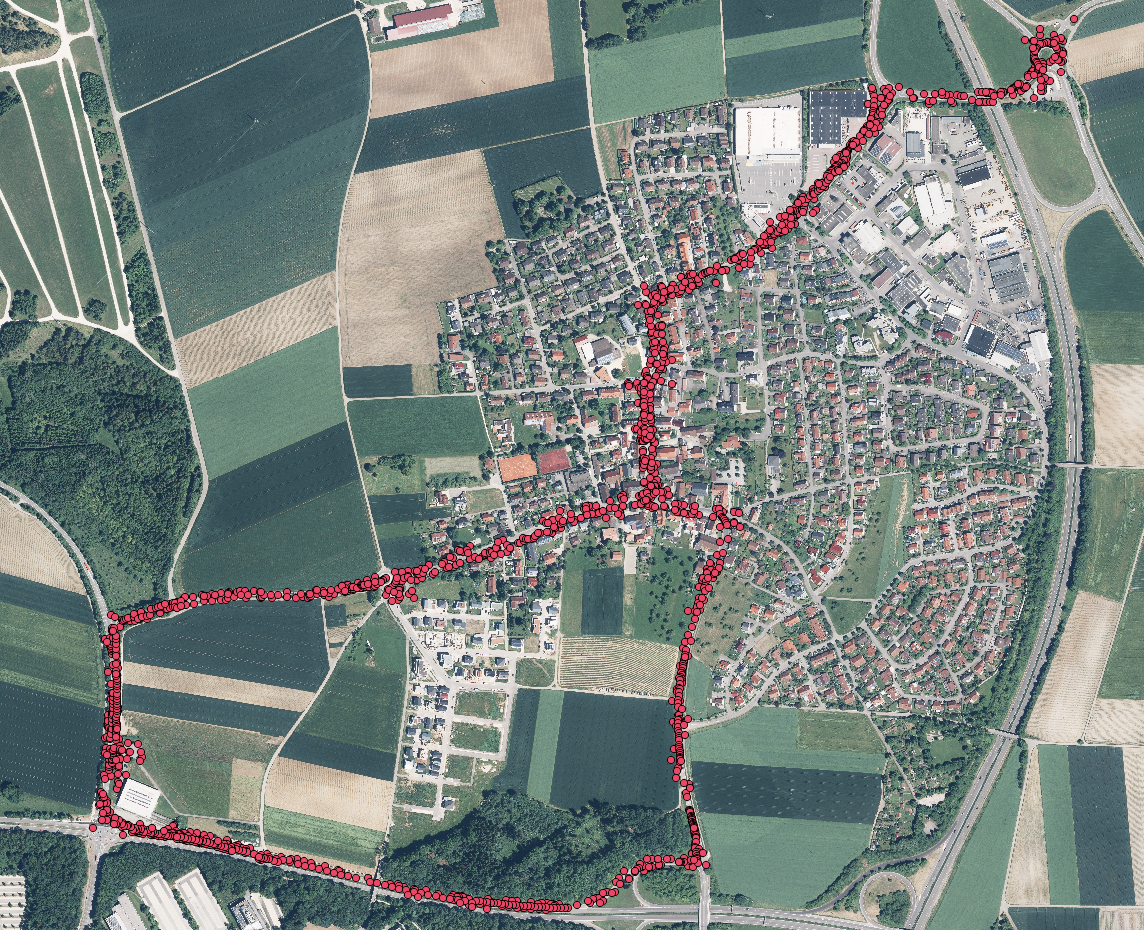}
    \caption{Overview of the landmark-based map with landmarks depicted as red circles (\textcolor{landmark_rgb}{$\bullet$}). The route in Ulm-Lehr (Germany) is about $5\,\si{\kilo\meter}$ long and the resulting map is comprised of 3860 landmarks.}
    \label{fig:landmark_map}
\end{figure}

\section{Ulm-Lehr Dataset and Map Creation}\label{sec:04_dataset}

We introduce our dataset to examine DeepLocalization in real-world scenarios. To that end, we have made recordings with our vehicle, equipped with a stereo camera setup, lidar and radar sensors, as well as a dGPS system for generating the ground-truth pose. Our test track in Ulm-Lehr, Germany, which is about $5\,\si{\kilo\meter}$ long, can be seen in Fig.~\ref{fig:landmark_map} along with the landmarks from the map.

The landmark-based map has been created in December 2017 by recording three runs on different days at our test track. It consists of $3860$ landmarks, of which $1731$ were obtained from lidar measurements, $1411$ from camera images and $718$ from radar targets. Independently, the measurement dataset has been built from eight runs on the Ulm-Lehr route in November 2018. In total, the dataset includes approximately $145.000$ samples, each sample containing measurements and the ground-truth pose. Six runs ($105.000$ samples) are used for training and validation. The remaining two runs ($40.000$ samples) are used for test. Below, we discuss the landmark extraction process and the generation of the landmark-based map in detail.

\begin{figure*}[t]
    \input{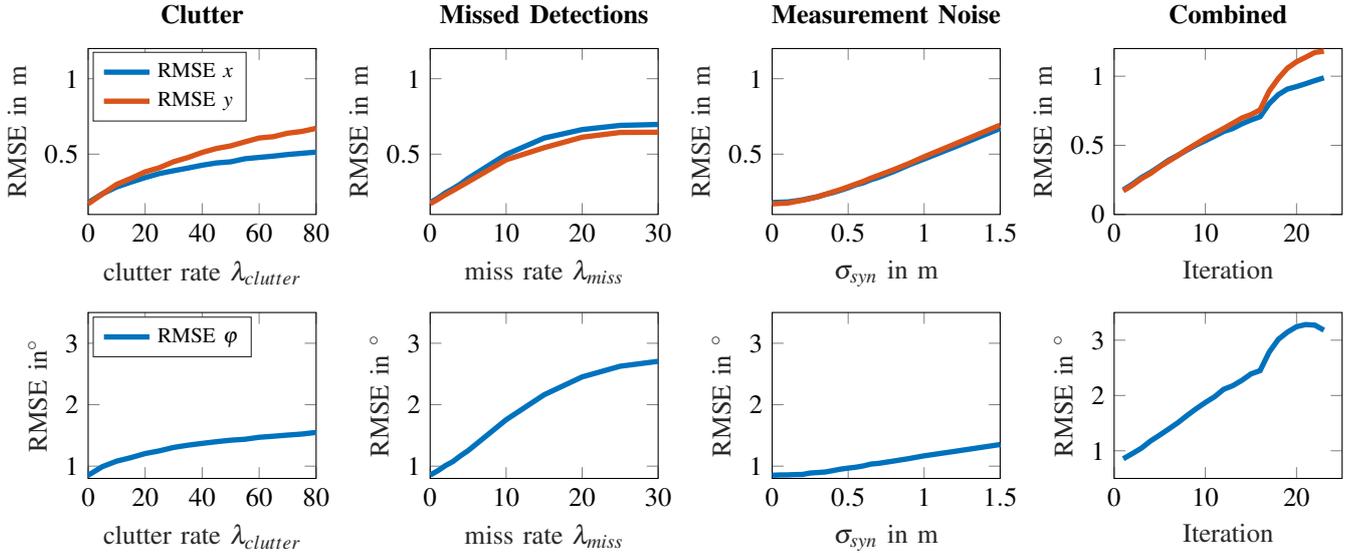}
    \caption{Overview of DeepLocalization results with synthetic measurements. The average amount of measurements per sample lies in the range of $20-30$. The first row depicts the RMSE results of the position and the bottom row shows the RMSE results of the orientation. In the first experiment, Poisson distributed clutter is added with $\lambda_{clutter} \in [0,80]$ (first column). Measurements are deleted in the second experiment with deletion rate $\lambda_{miss} \in [0,30]$ (second column). In the third experiment, noise is sampled randomly from a uniform distribution on the interval $[-\sigma_{syn}, \sigma_{syn}]$ with $\sigma_{syn} \in [0, 1.5]\,\si{\meter}$ (third column). The last experiment shows the result of the scenario, where all three effects occur simultaneously and are increased incrementally (fourth column). After iteration $18$, the impairments are too severe, causing the localization to diverge. }
    \label{fig:simulation_results}
\end{figure*}

\subsection{Landmark Extraction}\label{seq:measurement_dataset}
All measurement recordings take place with the same sensor configuration. While driving the route, we record measurements in the form of 3D point clouds for the lidar and radar sensors and RGB images with depth (RGB-D) from the stereo system. We extract features because we found them to be more stable and computational efficient than raw measurements. For the lidar and radar measurements, the density based DBSCAN \cite{ester1996density} cluster algorithm is applied to group adjacent points that have a minimum amount of neighboring points within a predefined distance. For the RGB-D images, maximally stable extremal regions (MSER) are extracted~\cite{matas2004robust}. The MSER algorithm is parameterized to detect road markings since they are permanent and thus can be detected recurrently. All features are stored in the vehicle coordinate system that has the origin at the center of the rear axle. The image features, in particular, are transformed to the vehicle coordinate system using the depth map. The extracted features from lidar, radar and RGB-D images define what we call landmarks. Additionally, the dGPS system records the ground-truth pose of the vehicle. It consists of the heading (yaw) and latitude and longitude coordinates that are transformed into Cartesian coordinates (UTM).

\subsection{Landmark-based Map}

The landmark extraction process (Sec.~\ref{seq:measurement_dataset}) is sufficient for extracting landmarks as measurements, but it is not enough for building our map. There are several landmark instances that are repeated several times. To avoid storing the same landmark multiple times, we additionally rely on the LMB-SLAM algorithm~\cite{deusch2015labeled} to make the landmark-based map sparser. In addition, dynamic objects, such as vehicles or time-dependent structures, may produce landmarks which are not valid for future use. To discard this type of landmarks, we fuse all three runs from December 2017 with a Bernoulli filtering approach \cite{stubler2017continuously}. Finally, the map landmarks are expressed in the UTM coordinate system. We have transformed the landmarks from the vehicle to the UTM coordinate system by incorporating the dGPS measurements. Therefore, the map landmarks have a relative accuracy of a few centimeters. In total, the map has a size of $600\,\si{\kilo\byte}$.

\section{Experiments}\label{sec:05_experiments}
We conduct two experiments on the Ulm-Lehr dataset to evaluate our approach. First, we examine our model with synthetic measurements and pose offsets. Second, we evaluate on the test set and compare our results with related approaches. For all experiments, we rely on the same DeepLocalization model that has about $1.8$~million learnable parameters. Our model has been trained with the ADAM optimizer~\cite{kingma2014adam} and a learning rate of $10^{-5}$. The batch size was set to $500$ samples. Additionally, all MLP networks have dropout~\cite{srivastava2014dropout} after every layer as regularizer, except the last one. During test, the inference time of our network is $1.758\,\si{\milli\second}$. Note for all evaluated approaches, including ours, the feature extraction phase is not included in the inference time. Our approach is implemented in \textit{\mbox{TensorFlow}}~\cite{tensorflow2015-whitepaper} and all operations are performed on a GPU with 11GB memory. 

\subsection{Synthetic Measurements Evaluation}\label{SynthEval}
We use synthetic measurements to examine the lower error bound of our approach. We also access how DeepLocalization generalizes under error-prone, missing or additional measurements.

For each time step, the map landmarks in the vehicle's field of view are used as measurements too. When the measurements and map landmarks are identical, the input is ideal and the only source of error can be the incorrect offset prediction of the network. To that end, we transform the measurements and map landmarks to the vehicle coordinate system. Then, a synthetic vehicle pose offset is added to the map landmarks as in the training process (Eq.~\eqref{eq:pinit}). To generate the vehicle pose offset, we sample noise from a uniform distribution on the interval $\sigma_{x}, \sigma_{y} \in [-2, 2]\,\si{\meter}$ and $\sigma_{\varphi} \in [-10, 10]\,^{\circ}$, which are realistic accuracies for modern GPS receivers~\cite{wing2005consumer}. We train our model based on the synthetic pose offsets. During test, our network receives the measurements and shifted map landmarks to predict the pose offset. Since the ground-truth is the synthetic vehicle pose offset, we can measure the prediction accuracy with the RMSE metric. We generate around $40.000$ test samples and achieve a RMSE of $0.178\,\si{\meter}$ and $0.17\,\si{\meter}$ for the $x$ and $y$ coordinates, respectively. Furthermore, the orientation error is $0.852\,^{\circ}$. The results are also summarized in Tab.~\ref{tab:real_world_results}.

Next, we consider the following situations: 
\begin{enumerate}
    \item Randomly delete measurements (simulating missed landmarks).
    \item Randomly add measurements (simulating clutter).
    \item Add randomly sampled noise to each measurement (simulating sensor noise).
    \item Add all three impairments simultaneously.
\end{enumerate}
 In practice, the rates at which measurements are added and deleted follow a Poisson distribution \mbox{$P_{\lambda}(k) = \lambda^{k}\exp(-\lambda)/(k!)$}, where $\lambda$ is the average number of events. The measurement noise for the third experiment is sampled from a uniform distribution on the interval $[-\sigma_{syn}, \sigma_{syn}]$. The evaluations are performed by increasing $\lambda$ and $\sigma_{syn}$. Our results are presented in Fig.~\ref{fig:simulation_results}.

One can observe that with clutter rates of $\lambda_{clutter} = 40$, the error remains low at $0.4\,\si{\meter}$, even though the average number of actual measurements is about $20-30$ per time step. Similar results are observed for missed detections with $\lambda_{miss} = 10$ and uniformly sampled noise for every landmark of about $\sigma_{syn} \in [-0.9, 0.9]\,\si{\meter}$. Finally, the robustness of our approach is demonstrated when all three impairments occur simultaneously. For instance, in iteration $10$ (\mbox{$\lambda_{clutter} = \lambda_{miss} = 10$} and \mbox{$\sigma_{syn}=0.27\,\si{\meter}$}) the RMSE value for $x$ and $y$ is about $0.5\,\si{\meter}$ and for the orientation $1.87^{\circ}$. When the effects become too severe (e.g. in iteration $18$), the measurements and the map landmarks diverge and, consequently, the localization fails.

\subsection{Ulm-Lehr Evaluation}

For this experiment, we train our network with the measurements and map landmarks from the Ulm-Lehr train set. To examine the robustness of our approach, we generate synthetic noise offsets to train three network models with different noise parameters. For the first setup, we sampled noise from a uniform distribution on the interval \mbox{$\sigma_{x} \in [-2, 2]\,\si{\meter}$}, \mbox{$\sigma_{y} \in [-2, 2]\,\si{\meter}$} and \mbox{$\sigma_{\varphi} \in [-10, 10]\,^{\circ}$}. For the second case, the noise is sampled on the interval \mbox{$\sigma_{x} \in [-1, 1]\,\si{\meter}$},~\mbox{$\sigma_{y} \in [-1, 1]\,\si{\meter}$} and \mbox{$\sigma_{\varphi} \in [-4, 4]\,^{\circ}$}; and the sampled noise for the last case was on the interval \mbox{$\sigma_{x} \in [-0.5, 0.5]\,\si{\meter}$}, \mbox{$\sigma_{y} \in [-0.5, 0.5]\,\si{\meter}$} and \mbox{$\sigma_{\varphi} \in [-2, 2]\,^{\circ}$}. In all three experiments we use GPS-based inference in order to improve the localization over the noisy GPS measurement. The results are summarized in Tab.~\ref{tab:real_world_results}.

Next, we evaluate the filter-based inference using the EKF. Here, we have relied on a single GPS measurement for initialization. Afterwards, the algorithm uses only the network output to obtain the current position, i.e. our prediction offset updates the pose from time step $t-1$ to $t$. Thus, after initialization, we are independent of any GPS measurements. After evaluating the three models, we have concluded that the network with noise parameters $\sigma_{x} = 1\,\si{\meter}$, \mbox{$\sigma_{y} = 1\,\si{\meter}$,~$ \sigma_{\varphi} = 4\,^{\circ}$} performs at best when combined with the EKF with a RMSE of $0.566\,\si{\meter}$ and $0.339\,\si{\meter}$ for \mbox{the $x$ and $y$ position}, respectively. Thus, we only report this result in Tab.~\ref{tab:real_world_results} for the EKF evaluation. The EKF is updated every $20~\si{\milli\second}$ and the measurement noise matrix was obtained empirically.

\begin{table}   
    \centering     
    \caption{Overview of results for synthetic measurements, real-world scenarios and related approaches.}
    \label{tab:real_world_results}
    \begin{tabular}{lcccc}
        \toprule
         & \multicolumn{3}{c}{\textbf{RMSE}}& inference \\
         \textbf{Experiment} & $x$ & $y$ & $\varphi$ & time \\
        \midrule
        \multicolumn{5}{l}{\textbf{Synthetic Measurements:}} \\[3pt] 
        DeepLocalization \\ $\sigma_{xy} = 2\,\si{\meter}, \sigma_{\varphi} = 10\,^{\circ}$ & $0.178\,\si{\meter}$ & $0.170\,\si{\meter}$ & $0.852\,^{\circ}$ & $1.758\,\si{\milli\second}$ \\[3pt]
        \midrule
        \multicolumn{5}{l}{\textbf{GPS-based Inference:}}\\[3pt]
        DeepLocalization \\ $\sigma_{xy} = 2\,\si{\meter}, \sigma_{\varphi} = 10\,^{\circ}$ & $0.773\,\si{\meter}$ & $0.701\,\si{\meter}$ & $2.52\,^{\circ}$ & $1.771\,\si{\milli\second}$ \\[3pt]
        DeepLocalization \\$\sigma_{xy} = 1\,\si{\meter}$,~$ \sigma_{\varphi} = 4\,^{\circ}$ & $0.447\,\si{\meter}$ & $0.377\,\si{\meter}$ & $1.37\,^{\circ}$ & $1.761\,\si{\milli\second}$ \\[3pt]
        DeepLocalization \\ $\sigma_{xy}=0.5\,\si{\meter}$,~$ \sigma_{\varphi} = 2\,^{\circ}$& $0.276\,\si{\meter}$ & $0.231\,\si{\meter}$ & $0.85\,^{\circ}$ & $1.728\,\si{\milli\second}$ \\[3pt]
        \midrule
        \multicolumn{5}{l}{\textbf{Filter-based Inference:}}\\[3pt]
        DeepLocalization \\+ EKF & $0.566\,\si{\meter}$ & $0.339\,\si{\meter}$ & $1.26\,^{\circ}$  & $2.133\,\si{\milli\second}$ \\[3pt]
        DeepLocalization  \\+ EKF + GPS & $0.271\,\si{\meter}$ & $0.245\,\si{\meter}$ & $0.82\,^{\circ}$ & $2.374\,\si{\milli\second}$ \\[3pt]
        \midrule
        \multicolumn{5}{l}{\textbf{Related Approaches:}} \\[3pt] 
        ICP \cite{besl1992method} & $1.174\,\si{\meter}$ & $1.467\,\si{\meter}$ & $4.96^{\circ}$ & $7.601\,\si{\milli\second}$ \\[3pt]
        EKF + GPS & $0.590\,\si{\meter}$ &$ 0.547\,\si{\meter}$ & $6.58^{\circ}$ & $0.168\,\si{\milli\second}$ \\[3pt]
        RFS-MCL + GPS \cite{stubler2017consistency} & $0.287\,\si{\meter}$ & $0.264\,\si{\meter}$ & $1.99\,^{\circ}$ & $28.92\,\si{\milli\second}$ \\[3pt]
        \bottomrule
     \end{tabular}
 \end{table}

In addition, we compare DeepLocalization with three approaches. We evaluate the iterative closest point algorithm (ICP)~\cite{besl1992method}, an extended Kalman Filter with only noisy GPS measurements as input and a random finite set Monte Carlo localization method (RFS-MCL)~\cite{stubler2017consistency}.

The ICP algorithm iteratively tries to minimize the difference between two point clouds, where one point cloud is fixed and the other one is transformed to match the reference point cloud. As initial guess, we transform the landmarks into the vehicle coordinate system using $H^{-1}$ with a noisy GPS measurement. Finally, the map landmarks are aligned to the measurements. Since the ICP algorithm is computationally expensive, the inference time is $7.601~\si{\milli\second}$ with accuracies (RMSE) of $1.1~\si{\meter}-1.4~\si{\meter}$ for the $x$ and $y$ coordinates and $5^{\circ}$ for the orientation.

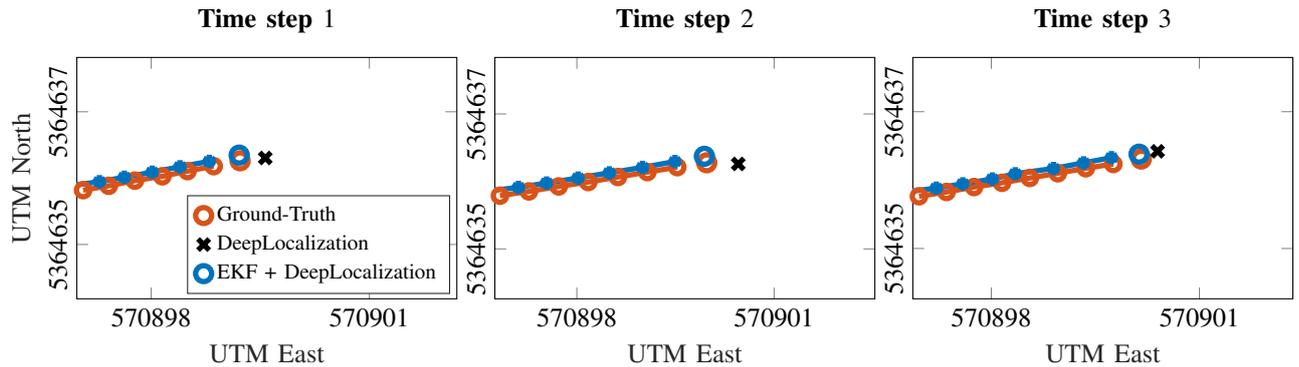
\begin{figure*}
    \centering
%
%
\definecolor{blue}{rgb}{0.00000,0.44700,0.74100}%
\definecolor{red}{rgb}{0.85000,0.32500,0.09800}%
\begin{tikzpicture}

\def\width{5}
\def\height{3.2}
\def\sep{0.5}

\def\firstrow{0}
\def\secondrow{-5.7}

\def\firstcol{0}
\def\secondcol{5.5}
\def\thirdcol{11}

\def\linewidth{2.0pt}
\def\marksingle{3.0pt}

\begin{axis}[%
width=\width cm,
height=\height cm,
at={(\firstcol cm, \firstrow cm)},
scale only axis,
xmin=-0.0774719460726497,
xmax=4.61881733668514,
xtick={0.841357619618066,3.53764690237585},
xticklabels={{570898},{570901}},
xlabel style={font=\color{white!15!black}},
xlabel={UTM East},
ymin=-2,
ymax=2.44265509955585,
ytick={-1,1.44265509955585},
yticklabels={{5364635},{5364637}},
yticklabel style={rotate=90},
ylabel style={font=\color{white!15!black}},
ylabel={UTM North},
axis background/.style={fill=white},
title style={font=\bfseries},
title={Time step $1$},
legend style={at={(0.29,0.02)}, anchor=south west,legend cell align=left, align=left, draw=white!15!black}
]
\addplot [color=red, line width=\linewidth, mark size=2.5pt, mark=o, mark options={solid, red}, forget plot]
  table[row sep=crcr]{%
0	0\\
0.316448943546432	0.07954313018665\\
0.639456175364210	0.17511244980365\\
0.97831126504895	0.2543541212017\\
1.28956465432132	0.350136454208\\
1.61345645646002	0.4416546422047\\
};
\addplot [color=blue, line width=\linewidth, mark size=2.5pt, mark=asterisk, mark options={solid, blue}, forget plot]
  table[row sep=crcr]{%
-0.158642380381934	0.0816516606137156\\
0.201917651109397	0.156102414242923\\
0.509640577831306	0.237890798598528\\
0.853979705367237	0.331526166759431\\
1.19644280383363	0.430737125687301\\
1.55906142550521	0.530342737212777\\
};
\addplot[only marks, mark=o, mark options={},line width=\linewidth, mark size=\marksingle, draw=red] table[row sep=crcr]{%
x	y\\
1.94023453788722211	0.5401435610869\\
};
\addlegendentry{\footnotesize Ground-Truth}

\addplot[only marks, mark=x, mark options={}, line width=\linewidth, mark size=\marksingle, draw=black] table[row sep=crcr]{%
x	y\\
2.251164690234575	0.592241564697822\\
};
\addlegendentry{\footnotesize DeepLocalization}

\addplot[only marks, mark=o, mark options={}, line width=\linewidth, mark size=\marksingle, draw=blue] table[row sep=crcr]{%
x	y\\
1.92894563528904	0.645677434956726\\
};
\addlegendentry{\footnotesize EKF + DeepLocalization}

\end{axis}

\begin{axis}[%
  width=\width cm,
  height=\height cm,
  at={(\secondcol cm, \firstrow cm)},
scale only axis,
xmin=-0.0559598524971394,
xmax=4.1026825278848,
xtick={0.841357619618066,3},
xticklabels={{570898},{570901}},
xlabel style={font=\color{white!15!black}},
xlabel={UTM East},
ymin=-1.93820639182522,
ymax=2.59744439325587,
ytick={-1,1.53565078508109},
yticklabels={{5364635},{5364637}},
yticklabel style={rotate=90},
axis background/.style={fill=white},
title style={font=\bfseries},
title={Time step $2$},
legend style={legend cell align=left, align=left, draw=white!15!black}
]
\addplot [color=red, line width=\linewidth, mark size=2.5pt, mark=o, mark options={solid, red}]
  table[row sep=crcr]{%
0	0\\
0.31645431214995	0.079564564546555\\
0.64245646546546	0.173565646546547\\
0.96601232154454	0.255545645623332\\
1.29112312125210	0.35014645664565\\
1.61345646456456	0.443056454565465\\
1.94031321233335	0.536014787850545\\
};

\addplot [color=blue, line width=\linewidth, mark size=2.5pt, mark=asterisk, mark options={solid, blue}]
  table[row sep=crcr]{%
-0.1614556456465	0.082065456467564\\
0.20215465446565	0.155874456556556\\
0.51000565465565	0.237890793234228\\
0.85407950547237	0.331423423425531\\
1.19658897897998	0.429437489398348\\
1.56042132133465	0.531213215672777\\
1.91599874321332	0.641876446341535\\
};

\addplot[only marks, mark=o, mark options={},line width=\linewidth, mark size=\marksingle, draw=red] table[row sep=crcr]{%
x	y\\
2.2624684766546	0.6180258158903\\
};

\addplot[only marks, mark=x, mark options={},line width=\linewidth, mark size=\marksingle,  draw=black] table[row sep=crcr]{%
x	y\\
2.60751490792911	0.596456456945721\\
};

\addplot[only marks, mark=o, mark options={},line width=\linewidth, mark size=\marksingle, draw=blue] table[row sep=crcr]{%
x	y\\
2.2371100213323	0.740132131658502\\
};

\end{axis}

\begin{axis}[%
  width=\width cm,
  height=\height cm,
  at={(\thirdcol cm, \firstrow cm)},
scale only axis,
xmin=-0.0712907004880109,
xmax=4.3508881141653,
xtick={0.841357619618066,3.26353643427137},
xticklabels={{570898},{570901}},
xlabel style={font=\color{white!15!black}},
xlabel={UTM East},
ymin=-1.96225451325302,
ymax=2.65527674256588,
ytick={-1,1.6175312558189},
yticklabels={{5364635},{5364637}},
yticklabel style={rotate=90},
axis background/.style={fill=white},
title style={font=\bfseries},
title={Time step $3$},
legend style={legend cell align=left, align=left, draw=white!15!black}
]
\addplot [color=red, line width=\linewidth, mark size=2.5pt, mark=o, mark options={solid, red}]
  table[row sep=crcr]{%
0	0\\
0.320234578348795	0.080987123122355\\
0.640343543769426	0.17588152473193\\
0.96612355607252	0.257063779781311\\
1.29031901189592	0.349659430794418\\
1.61589354505143	0.439676550998185\\
1.94137363374113	0.535653450813404\\
2.26122397033235	0.616789223138903\\
};

\addplot [color=blue, line width=\linewidth, mark size=2.5pt, mark=asterisk, mark options={solid, blue}]
  table[row sep=crcr]{%
-0.15611480381934	0.082111266012756\\
0.202234176123007	0.1551239814270143\\
0.51004564831336	0.238145607987358\\
0.85411355367631	0.330345686711768\\
1.12034554338963	0.43063455611304\\
1.565456777234521	0.53189635456379\\
1.91587625852099	0.63901327603361\\
2.24009823292109	0.73834560352855\\
};

\addplot[only marks, mark=o, mark options={}, line width=\linewidth, mark size=\marksingle, draw=red] table[row sep=crcr]{%
x	y\\
2.5864564565492	0.7112852356413\\
};

\addplot[only marks, mark=x, mark options={}, line width=\linewidth, mark size=\marksingle, draw=black] table[row sep=crcr]{%
x	y\\
2.77629825344291	0.859734241123443\\
};

\addplot[only marks, mark=o, mark options={}, line width=\linewidth, mark size=\marksingle, draw=blue] table[row sep=crcr]{%
x	y\\
2.561345345434	0.80546367567675\\
};

\end{axis}

\end{tikzpicture}%
    \caption{Exemplary sequence of DeepLocalization results in combination with the  EKF. We show three consecutive samples with the preceding ground-truth and predicted trajectory. The corrected pose from DeepLocalization is depicted as a black cross and the smoothed output of the EKF as a blue circle. Although the network was trained with synthetic noise offsets, it generalizes to real-world applications and corrects the pose from the previous result. Additionally, small errors are adjusted by the EKF with an underlying CTRV motion model.}
    \label{fig:sequence_plots}
\end{figure*}

The EKF with only noisy GPS measurements as input achieves slightly better results for the position accuracy which are reported in Tab.~\ref{tab:real_world_results}. The Monte-Carlo localization method uses the same map and features from the multi-modal measurements. Unlike our learning-based approach, \etal{Stuebler}~model the measurements and landmarks as Random Finite Sets (RFS), and estimate the vehicle's pose with a particle filter. The weights of the particles are updated using a Bernoulli filtering approach. An in-depth explanation of the RFS-MCL method can be found in~\cite{stubler2017consistency}. We have re-implemented the approach with a total of $1000$ particles which translates to $28.92\,\si{\milli\second}$ of inference time. In our implementation, the method uses additional GPS measurements. For that reason, we also evaluate results by incorporating GPS measurements in the EKF filter, i.e. DeepLocalization + EKF + GPS in Tab.~\ref{tab:real_world_results}. We achieve position RMSE in the range of $0.245\,\si{\meter} - 0.271\,\si{\meter}$ and an orientation error of $0.82^{\circ}$. Our approach has a computational time of $2.374\,\si{\milli\second}$, which is about ten times faster than the traditional RFS-MCL. Finally, a visual illustration of DeepLocalization with an EKF is shown in Fig. \ref{fig:sequence_plots}. It becomes apparent that our prediction is very close to the ground-truth position in the UTM coordinate system.

\section{Conclusion}\label{sec:06_conclusion}
In this work, we have proposed a new approach for vehicle self-localization. We have introduced DeepLocalization, a deep neural network, that regresses the translation and rotation offsets from unordered and dynamic input lists. We have presented two inference algorithms, namely GPS-based and filter-based inference. In the evaluation, we have shown better performance than the prior work on the Ulm-Lehr dataset. In addition, our approach runs ten times faster than the RFS-MCL. As future work, we would like to explore how to be completely independent of GPS measurements. 



\begin{thebibliography}{10}
       \providecommand{\url}[1]{#1}
       \csname url@rmstyle\endcsname
       \providecommand{\newblock}{\relax}
       \providecommand{\bibinfo}[2]{#2}
       \providecommand\BIBentrySTDinterwordspacing{\spaceskip=0pt\relax}
       \providecommand\BIBentryALTinterwordstretchfactor{4}
       \providecommand\BIBentryALTinterwordspacing{\spaceskip=\fontdimen2\font plus
       \BIBentryALTinterwordstretchfactor\fontdimen3\font minus
         \fontdimen4\font\relax}
       \providecommand\BIBforeignlanguage[2]{{%
       \expandafter\ifx\csname l@#1\endcsname\relax
       \typeout{** WARNING: IEEEtran.bst: No hyphenation pattern has been}%
       \typeout{** loaded for the language `#1'. Using the pattern for}%
       \typeout{** the default language instead.}%
       \else
       \language=\csname l@#1\endcsname
       \fi
       #2}}
       
       \bibitem{thrun2005probabilistic}
       S.~Thrun, W.~Burgard, and D.~Fox, \emph{Probabilistic robotics}.\hskip 1em plus
         0.5em minus 0.4em\relax MIT press, 2005.
       
       \bibitem{kunz2015autonomous}
       F.~Kunz, \emph{et~al.}, ``Autonomous driving at {Ulm} university: A modular,
         robust, and sensor-independent fusion approach,'' in \emph{2015 IEEE
         intelligent vehicles symposium (IV)}.\hskip 1em plus 0.5em minus 0.4em\relax
         IEEE, 2015, pp. 666--673.
       
       \bibitem{gies2018environment}
       F.~Gies, A.~Danzer, and K.~Dietmayer, ``Environment perception framework fusing
         multi-object tracking, dynamic occupancy grid maps and digital maps,'' in
         \emph{2018 21st International Conference on Intelligent Transportation
         Systems (ITSC)}.\hskip 1em plus 0.5em minus 0.4em\relax IEEE, 2018, pp.
         3859--3865.
       
       \bibitem{mohamed1999adaptive}
       A.~Mohamed and K.~Schwarz, ``Adaptive {Kalman} filtering for {INS/GPS},''
         \emph{Journal of geodesy}, vol.~73, no.~4, pp. 193--203, 1999.
       
       \bibitem{kos2010effects}
       T.~Kos, I.~Markezic, and J.~Pokrajcic, ``Effects of multipath reception on
         {GPS} positioning performance,'' in \emph{Proceedings ELMAR-2010}.\hskip 1em
         plus 0.5em minus 0.4em\relax IEEE, 2010, pp. 399--402.
       
       \bibitem{cummins2008fab}
       M.~Cummins and P.~Newman, ``{FAB-MAP}: Probabilistic localization and mapping
         in the space of appearance,'' \emph{The International Journal of Robotics
         Research}, vol.~27, no.~6, pp. 647--665, 2008.
       
       \bibitem{ziegler2014video}
       J.~Ziegler, \emph{et~al.}, ``Video based localization for {Bertha},'' in
         \emph{Intelligent Vehicles Symposium Proceedings, 2014 IEEE}.\hskip 1em plus
         0.5em minus 0.4em\relax IEEE, 2014, pp. 1231--1238.
       
       \bibitem{brubaker2016map}
       M.~A. Brubaker, A.~Geiger, and R.~Urtasun, ``Map-based probabilistic visual
         self-localization,'' \emph{IEEE transactions on pattern analysis and machine
         intelligence}, vol.~38, no.~4, pp. 652--665, 2016.
       
       \bibitem{wiest2014localization}
       J.~Wiest, \emph{et~al.}, ``Localization based on region descriptors in grid
         maps,'' in \emph{2014 IEEE Intelligent Vehicles Symposium Proceedings}.\hskip
         1em plus 0.5em minus 0.4em\relax IEEE, 2014, pp. 793--799.
       
       \bibitem{deusch2015labeled}
       H.~Deusch, S.~Reuter, and K.~Dietmayer, ``The labeled multi-bernoulli {SLAM}
         filter,'' \emph{IEEE Signal Processing Letters}, vol.~22, no.~10, pp.
         1561--1565, 2015.
       
       \bibitem{stubler2017consistency}
       M.~St{\"u}bler, S.~Reuter, and K.~Dietmayer, ``Consistency of feature-based
         random-set {Monte-Carlo} localization,'' in \emph{2017 European Conference on
         Mobile Robots (ECMR)}.\hskip 1em plus 0.5em minus 0.4em\relax IEEE, 2017, pp.
         1--6.
       
       \bibitem{qi2017pointnet}
       C.~R. Qi, H.~Su, K.~Mo, and L.~J. Guibas, ``{PointNet}: Deep learning on point
         sets for {3D} classification and segmentation,'' \emph{Proc. Computer Vision
         and Pattern Recognition (CVPR), IEEE}, vol.~1, no.~2, p.~4, 2017.
       
       \bibitem{dellaert1999monte}
       F.~Dellaert, D.~Fox, W.~Burgard, and S.~Thrun, ``Monte {Carlo} localization for
         mobile robots,'' in \emph{ICRA}, vol.~2, 1999, pp. 1322--1328.
       
       \bibitem{thrun2001robust}
       S.~Thrun, D.~Fox, W.~Burgard, and F.~Dellaert, ``Robust {Monte Carlo}
         localization for mobile robots,'' \emph{Artificial intelligence}, vol. 128,
         no. 1-2, pp. 99--141, 2001.
       
       \bibitem{levinson2007map}
       J.~Levinson, M.~Montemerlo, and S.~Thrun, ``Map-based precision vehicle
         localization in urban environments.'' in \emph{Robotics: Science and
         Systems}, vol.~4.\hskip 1em plus 0.5em minus 0.4em\relax Citeseer, 2007,
         p.~1.
       
       \bibitem{levinson2010robust}
       J.~Levinson and S.~Thrun, ``Robust vehicle localization in urban environments
         using probabilistic maps,'' in \emph{2010 IEEE International Conference on
         Robotics and Automation}.\hskip 1em plus 0.5em minus 0.4em\relax IEEE, 2010,
         pp. 4372--4378.
       
       \bibitem{wolcott2015fast}
       R.~W. Wolcott and R.~M. Eustice, ``Fast lidar localization using
         multiresolution gaussian mixture maps,'' in \emph{Robotics and Automation
         (ICRA), 2015 IEEE International Conference on}.\hskip 1em plus 0.5em minus
         0.4em\relax IEEE, 2015, pp. 2814--2821.
       
       \bibitem{besl1992method}
       P.~J. Besl and N.~D. McKay, ``Method for registration of {3-D} shapes,'' in
         \emph{Sensor Fusion IV: Control Paradigms and Data Structures}, vol.
         1611.\hskip 1em plus 0.5em minus 0.4em\relax International Society for Optics
         and Photonics, 1992, pp. 586--607.
       
       \bibitem{segal2009generalized}
       A.~Segal, D.~Haehnel, and S.~Thrun, ``Generalized-{ICP}.'' in \emph{Robotics:
         science and systems}, vol.~2, no.~4, 2009, p. 435.
       
       \bibitem{kummerle2008monte}
       R.~K{\"u}mmerle, R.~Triebel, P.~Pfaff, and W.~Burgard, ``{Monte Carlo}
         localization in outdoor terrains using multilevel surface maps,''
         \emph{Journal of Field Robotics}, vol.~25, no. 6-7, pp. 346--359, 2008.
       
       \bibitem{belagiannis2015robust}
       V.~Belagiannis, C.~Rupprecht, G.~Carneiro, and N.~Navab, ``Robust optimization
         for deep regression,'' in \emph{Proceedings of the IEEE international
         conference on computer vision}, 2015, pp. 2830--2838.
       
       \bibitem{kendall2015posenet}
       A.~Kendall, M.~Grimes, and R.~Cipolla, ``{PoseNet}: A convolutional network for
         real-time 6-{DOF} camera relocalization,'' in \emph{Proceedings of the IEEE
         international conference on computer vision}, 2015, pp. 2938--2946.
       
       \bibitem{valada2018deep}
       A.~Valada, N.~Radwan, and W.~Burgard, ``Deep auxiliary learning for visual
         localization and odometry,'' in \emph{2018 IEEE International Conference on
         Robotics and Automation (ICRA)}.\hskip 1em plus 0.5em minus 0.4em\relax IEEE,
         2018, pp. 6939--6946.
       
       \bibitem{ding2018deepmapping}
       L.~Ding and C.~Feng, ``{DeepMapping}: Unsupervised map estimation from multiple
         point clouds,'' \emph{arXiv preprint arXiv:1811.11397}, 2018.
       
       \bibitem{kendall2018multi}
       A.~Kendall, Y.~Gal, and R.~Cipolla, ``Multi-task learning using uncertainty to
         weigh losses for scene geometry and semantics,'' in \emph{Proceedings of the
         IEEE Conference on Computer Vision and Pattern Recognition}, 2018, pp.
         7482--7491.
       
       \bibitem{kendall2017uncertainties}
       A.~Kendall and Y.~Gal, ``What uncertainties do we need in bayesian deep
         learning for computer vision?'' in \emph{Advances in neural information
         processing systems}, 2017, pp. 5574--5584.
       
       \bibitem{kendall2017geometric}
       A.~Kendall, R.~Cipolla, \emph{et~al.}, ``Geometric loss functions for camera
         pose regression with deep learning,'' in \emph{Proc. CVPR}, vol.~3, 2017,
         p.~8.
       
       \bibitem{ester1996density}
       M.~Ester, H.-P. Kriegel, J.~Sander, X.~Xu, \emph{et~al.}, ``A density-based
         algorithm for discovering clusters in large spatial databases with noise.''
         in \emph{Kdd}, vol.~96, no.~34, 1996, pp. 226--231.
       
       \bibitem{matas2004robust}
       J.~Matas, O.~Chum, M.~Urban, and T.~Pajdla, ``Robust wide-baseline stereo from
         maximally stable extremal regions,'' \emph{Image and vision computing},
         vol.~22, no.~10, pp. 761--767, 2004.
       
       \bibitem{stubler2017continuously}
       M.~St{\"u}bler, S.~Reuter, and K.~Dietmayer, ``A continuously learning
         feature-based map using a bernoulli filtering approach,'' in \emph{2017
         Sensor Data Fusion: Trends, Solutions, Applications (SDF)}.\hskip 1em plus
         0.5em minus 0.4em\relax IEEE, 2017, pp. 1--6.
       
       \bibitem{kingma2014adam}
       D.~P. Kingma and J.~Ba, ``{Adam}: A method for stochastic optimization,''
         \emph{arXiv preprint arXiv:1412.6980}, 2014.
       
       \bibitem{srivastava2014dropout}
       N.~Srivastava, G.~Hinton, A.~Krizhevsky, I.~Sutskever, and R.~Salakhutdinov,
         ``Dropout: a simple way to prevent neural networks from overfitting,''
         \emph{The Journal of Machine Learning Research}, vol.~15, no.~1, pp.
         1929--1958, 2014.
       
       \bibitem{tensorflow2015-whitepaper}
       \BIBentryALTinterwordspacing
       M.~Abadi, \emph{et~al.}, ``{TensorFlow}: Large-scale machine learning on
         heterogeneous systems,'' 2015, software available from tensorflow.org.
         [Online]. Available: \url{https://www.tensorflow.org/}
       \BIBentrySTDinterwordspacing
       
       \bibitem{wing2005consumer}
       M.~G. Wing, A.~Eklund, and L.~D. Kellogg, ``Consumer-grade global positioning
         system ({GPS}) accuracy and reliability,'' \emph{Journal of forestry}, vol.
         103, no.~4, pp. 169--173, 2005.
       
\end{thebibliography}
\end{document}